\documentclass[11pt]{article}
\usepackage{acl2005}
\usepackage{times}
\usepackage{psfig}
\usepackage{haldefs}
\usepackage{algorithm}
\usepackage{algorithmic}
\usepackage{url}
\newcommand{\mention}[3]{\underline{#1}${}_{\textsc{#2}}^{\textsc{#3}}$}

\title{A Large-Scale Exploration of Effective Global Features\\%
       for a Joint Entity Detection and Tracking Model}

\author{Hal Daum\'e III \textnormal{and} Daniel Marcu\\
Information Sciences Institute\\
4676 Admiralty Way, Suite 1001\\
Marina del Rey, CA 90292\\
\texttt{\{hdaume,marcu\}@isi.edu}\\
}

\date{}

\begin{document}

\maketitle

\begin{abstract}
Entity detection and tracking (EDT) is the task of identifying textual
mentions of real-world entities in documents, extending the named
entity detection and coreference resolution task by considering
mentions other than names (pronouns, definite descriptions, etc.).
Like NE tagging and coreference resolution, most solutions to the EDT
task separate out the mention detection aspect from the coreference
aspect.  By doing so, these solutions are limited to using only local
features for learning.  In contrast, by modeling both aspects of the
EDT task simultaneously, we are able to learn using highly complex,
non-local features.  We develop a new joint EDT model and explore the
utility of many features, demonstrating their effectiveness on this
task.
\end{abstract}

\vspace{-1.5mm}\section{Introduction}\vspace{-1.5mm}

In many natural language applications, such as automatic document
summarization, machine translation, question answering and information
retrieval, it is advantageous to pre-process text documents to
identify references to entities.  An entity, loosely defined, is a
person, location, organization or geo-political entity (GPE) that
exists in the real world.  Being able to identify references to
real-world entities of these types is an important and difficult
natural language processing problem.  It involves finding text spans
that correspond to an entity, identifying what \emph{type of entity}
it is (person, location, etc.), identifying what \emph{type of
mention} it is (name, nominal, pronoun, etc.) and finally identifying
which \emph{other} mentions in the document it corefers with.  The
difficulty lies in the fact that there are often many ambiguous ways
to refer to the same entity.  For example, consider the two sentences
below:

\begin{quote} \small
\mention{Bill Clinton}{per--1}{nam} gave a speech today to
\mention{the Senate}{org--2}{nam} .  \mention{The
President}{per--1}{nom} outlined \mention{his}{per--1}{pro} plan for
budget reform to \mention{them}{org--2}{pro} .
\end{quote}

There are five entity \emph{mentions} in these two sentences, each of
which is underlined (the corresponding mention type and entity type
appear as superscripts and subscripts, respectively, with coreference
chains marked in the subscripts), but only two \emph{entities}: \{
Bill Clinton, The president, his \} and \{ the Senate, them \}.  The
\emph{mention detection} task is to identify the entity mentions and
their types, without regard for the underlying entity sets, while
\emph{coreference resolution} groups a given mentions into sets.

Current state-of-the-art solutions to this problem split it into two
parts: mention detection and coreference
\cite{soon01machine,ng-cardie-02,florian04edt}.  First, a model is run
that attempts to identify each mention in a text and assign it a type
(person, organization, etc.).  Then, one holds these mentions fixed
and attempts to identify which ones refer to the same entity.  This is
typically accomplished through some form of clustering, with
clustering weights often tuned through some local learning procedure.
This pipelining scheme has the significant drawback that the mention
detection module cannot take advantage of information from the
coreference module.  Moreover, within the coreference task, performing
learning and clustering as separate tasks makes learning rather
ad-hoc.

In this paper, we build a model that solves the mention detection and
coreference problems in a simultaneous, joint manner.  By doing so, we
are able to obtain an empirically superior system as well as integrate
a large collection of features that one cannot consider in the
standard pipelined approach.  Our ability to perform this modeling is
based on the \emph{Learning as Search Optimization} framework, which
we review in Section~\ref{sec:LaSO}.  In Section~\ref{sec:model}, we
describe our joint EDT model in terms of the search procedure
executed.  In Section~\ref{sec:features}, we describe the features we
employ in this model; these include the standard lexical, semantic
(WordNet) and string matching features found in most other systems.
We additionally consider many other feature types, most interestingly
\emph{count-based features}, which take into account the distribution
of entities and mentions (and are not expressible in the binary
classification method for coreference) and \emph{knowledge-based
features}, which exploit large corpora for learning name-to-nominal
references.  In Section~\ref{sec:results}, we present our experimental
results.  First, we compare our joint system with a pipelined version
of the system, and show that joint inference leads to improved
performance.  Next, we perform an extensive feature comparison
experiment to determine which features are most useful for the
coreference task, showing that our newly introduced features provide
useful new information.  We conclude in Section~\ref{sec:discussion}.

\vspace{-1.5mm}\section{Learning as Search Optimization} \label{sec:LaSO}\vspace{-1.5mm}


When one attempts to apply current, standard machine learning
algorithms to problems with combinatorial structured outputs, the
resulting algorithm implicitly assumes that it is possible to find the
best structures for a given input (and some model parameters).
Furthermore, most models require much more, either in the form of
feature expectations for conditional likelihood-based methods
\cite{lafferty01crf} or local marginal distributions for margin-based
methods \cite{taskar03mmmn}.  In many cases---including EDT and
coreference---this is a false assumption.  Often, we are not able to
find the \emph{best} solution, but rather must employ an approximate
search to find the best possible solution, given time and space
constraints.  The Learning as Search Optimization (LaSO) framework
exploits this difficulty as an opportunity and seeks to find model
parameters that are good \emph{within the context of search.}

\begin{figure}[t]
\footnotesize
\center
\framebox{\hspace{-2mm}\begin{minipage}[t]{7cm}
\begin{algorithmic}
\STATE {\bf Algo} Learn(\textit{problem}, \textit{initial}, \textit{enqueue}, $\vec w$, $x$, $y$)
\STATE \textit{nodes} $\leftarrow$ MakeQueue(MakeNode(\textit{problem},\textit{initial}))
\WHILE{\textit{nodes} is not empty}
\STATE \textit{node} $\leftarrow$ RemoveFront(\textit{nodes})
\IF{none of $\textit{nodes} \cup \{\textit{node}\}$ is $y$-good {\bf
    or}\\ ~~~~GoalTest(\textit{node}) and \textit{node} is not $y$-good}
\STATE \textit{sibs}  $\leftarrow$ \textit{siblings}(\textit{node}, $y$)
\STATE $\vec w \leftarrow$ \textit{update}($\vec w$, $x$, \textit{sibs}, \textit{node} $\cup$ \textit{nodes})
\STATE \textit{nodes} $\leftarrow$ MakeQueue(\textit{sibs})
\ELSE
\STATE {\bf if} GoalTest(\textit{node}) {\bf then return} $\vec w$
\STATE \textit{next} $\leftarrow$ Operators(\textit{node})
\STATE \textit{nodes} $\leftarrow$ \textit{enqueue}(\textit{problem}, \textit{nodes}, \textit{next}, $\vec w$)
\ENDIF
\ENDWHILE
\end{algorithmic}
\end{minipage}\hspace{2mm}}
\caption{The generic search/learning algorithm.}
\label{fig:learn}
\vspace{-5mm}
\end{figure}

More formally, following the LaSO framework, we assume that there is a
set of input structures $\cX$ and a set of output structures $\cY$ (in
our case, elements $x \in \cX$ will be documents and elements $y \in
\cY$ will be documents marked up with mentions and their coreference
sets).  Additionally, we provide the structure of a search space $\cS$
that results in elements of $\cY$ (we will discuss our choice for this
component later in Section~\ref{sec:model}).  The LaSO framework
relies on a \emph{monotonicity} assumption: given a structure $y \in
\cY$ and a node $n$ in the search space, we must be able to calculate
whether it is \emph{possible} for this node $n$ to eventually lead to
$y$ (such nodes are called $y$-good).

LaSO parameterizes the search process with a weight vector $\vec w \in
\R^D$, where weights correspond to features of search space nodes and
inputs.  Specifically, we write $\Ph : \cX \times \cS \fto \R^D$ as a
function that takes a pair of an input $x$ and a node in the search
space $n$ and produces a vector of features.  LaSO takes a standard
search algorithm and modifies it to incorporate learning in an online
manner to the algorithm shown in Figure~\ref{fig:learn}.  The key idea
is to perform search as normal until a point at which it becomes
impossible to reach the correct solution.  When this happens, the
weight vector $\vec w$ is updated in a corrective fashion.  The
algorithm relies on a parameter update formula; the two suggested by
\cite{daume05laso} are a standard Perceptron-style update and an
approximate large margin update of the sort proposed by
\cite{gentile-alma}.  In this work, we only use the large margin
update, since in the original LaSO work, it consistently outperformed
the simpler Perceptron updates.  The update has the form given below:

\begin{small}\vspace{-4mm}
\begin{align*}
\vec w & \leftarrow \textit{proj}~(\vec w + C k^{-1/2} ~ \vec \De) \\
\vec \De & = 
\textit{proj}\left[
                \sum_{n \in \textit{sibs }} \frac {\vec \Ph(x, n)} {|\textit{sibs}|} - 
                \sum_{n \in \textit{nodes}} \frac {\vec \Ph(x, n)} {|\textit{nodes}|}\right]
\end{align*}\vspace{-4mm}
\end{small}

\noindent
Where $k$ is the update number, $C$ is a tunable parameter and
$\textit{proj}$ projects a vector into the unit sphere.

\vspace{-1.5mm}\section{Joint EDT Model} \label{sec:model}\vspace{-1.5mm}

The LaSO framework essentially requires us to specify two components:
the search space (and corresponding operations) and the features.
These two are inherently tied, since the features rely on the search
space, but for the time being we will ignore the issue of the feature
functions and focus on the search.

\vspace{-1.5mm}
\subsection{Search Space}

We structure search in a left-to-right decoding framework: a
hypothesis is a complete identification of the initial segment of a
document.  For instance, on a document with $N$ words, a hypothesis
that ends at position $0 < n < N$ is essentially what you would get if
you took the full structured output and chopped it off at word $n$.
In the example given in the introduction, one hypothesis might
correspond to ``\underline{Bill Clinton} gave a'' (which would be a
$y$-good hypothesis), or to ``\underline{Bill} Clinton \underline{gave
a}'' (which would not be a $y$-good hypothesis).

A hypothesis is expanded through the application of the search
operations.  In our case, the search procedure first chooses the
number of words it is going to consume (for instance, to form the
mention ``Bill Clinton,'' it would need to consume two words).  Then,
it decides on an entity type and a mention type (or it opts to call
this chunk not an entity (NAE), corresponding to non-underlined
words).  Finally, assuming it did not choose to form an NAE, it
decides on which of the foregoing coreference chains this entity
belongs to, or none (if it is the first mention of a new entity).  All
these decisions are made simultaneously, and the given hypothesis is
then scored.

\vspace{-1.5mm}
\subsection{An Example}

For concreteness, consider again the text given in the introduction.
Suppose that we are at the word ``them'' and the hypothesis we are
expanding is correct.  That is, we have correctly identified ``Bill
Clinton'' with entity type ``person'' and mention type ``name;'' that
we have identified ``the Senate'' with entity type ``organization''
and mention type ``name;'' and that we have identified both ``The
President'' and ``his'' as entities with entity type ``person'' and
mention types ``nominal'' and ``pronoun,'' respectively, and that
``The President'' points back to the chain $\langle$Bill
Clinton$\rangle$ and that ``his'' points back to the chain
$\langle$Bill Clinton, The President$\rangle$.

At this point of search, we have two choices for length: one or two
(because there are only two words left: ``them'' and a period).  A
first hypothesis would be that the word ``them'' is NAE.  A second
hypothesis would be that ``them'' is a named person and is a new
entity; a third hypothesis would be that ``them'' is a named person
and is coreference with the ``Bill Clinton'' chain; a fourth
hypothesis would be that ``them'' is a pronominal organization and is
a new entity; next, ``them'' could be a pronominal organization that
is coreferent with ``the Senate''; and so on.  Similar choices would
be considered for the string ``them .'' when two words are selected.



\vspace{-1.5mm}
\subsection{Linkage Type}

One significant issue that arises in the context of assigning a
hypothesis to a coreference chain is how to compute features over that
chain.  As we will discuss in Section~\ref{sec:features}, the majority
of our coreference-specific features are over \emph{pairs} of chunks:
the proposed new mention and an antecedent.  However, since in general
a proposed mention can have well more than one antecedent, we are left
with a decision about how to combine this information.

The first, most obvious solution, is to essentially do nothing: simply
compute the features over all pairs and add them up as usual.  This
method, however, intuitively has the potential for over-counting the
effects of large chains.  To compensate for this, one might advocate
the use of an \emph{average link} computation, where the score for a
coreference chain is computed by averaging over its elements.  One
might also consider a \emph{max link} or \emph{min link} scenario,
where one of the extrema is chosen as the value.  Other research has
suggested that a simple \emph{last link}, where a mention is simply
matched against the most recent mention in a chain might be
appropriate, while \emph{first link} might also be appropriate because
the first mention of an entity tends to carry the most information.

In addition to these standard linkages, we also consider an
\emph{intelligent link} scenario, where the method of computing the
link structure depends on the \emph{mention type}.  The intelligent
link is computed as follow, based on the mention type of the current
mention, $m$:

\begin{description}
\vspace{-2mm}
\item[If $m=$NAM then:] match \emph{first} on NAM elements in the
  chain; if there are none, match against the \emph{last} NOM element;
  otherwise, use \emph{max link}.

\vspace{-2mm}
\item[If $m=$NOM then:] match against the \emph{max} NOM in the chain;
  otherwise, match against the most \emph{last} NAM; otherwise, use
  \emph{max link}.

\vspace{-2mm}
\item[If $m=$PRO then:] use \emph{average link} across all PRO or NAM;
if there are none, use \emph{max link}.
\end{description}
\vspace{-2mm}

The construction of this methodology as guided by intuition (for
instance, matching names against names is easy, and the first name
tends to be the most complete) and subsequently tuned by
experimentation on the development data.  One might consider
\emph{learning} the best link method, and this may result in better
performance, but we do not explore this option in this work.  The
initial results we present will be based on using intelligent link,
but we will also compare the different linkage types explicitly.

\vspace{-1.5mm}\section{Feature Functions} \label{sec:features}\vspace{-1.5mm}

All the features we consider are of the form \textit{base-feature}
$\times$ \textit{decision-feature}, where base features are functions
of the input and decisions are functions of the hypothesis.  For
instance, a base feature might be something like ``the current chunk
contains the word 'Clinton''' and a decision feature might be
something like ``the current chunk is a named person.''

\vspace{-1.5mm}
\subsection{Base Features}

For pedagogical purposes and to facility model comparisons, we have
separated the base features into eleven classes: lexical, syntactic,
pattern-based, count-based, semantic, knowledge-based, class-based,
list-based, inference-based, string match features and history-based
features.  We will deal with each of these in turn.  Finally, we will
discuss how these base features are combined into \emph{meta-features}
that are actually used for prediction.

\paragraph{Lexical features.}

The class of lexical features contains simply computable features of
single words.  This includes: the number of words in the current
chunk; the unigrams (words) contained in this chunk; the bigrams; the
two character prefixes and suffixes; the word stem; the case of the
word, computed by regular expressions like those given by
\cite{bikel99name}; simple morphological features (number, person and
tense when applicable); and, in the case of coreference, pairs of
features between the current mention and an antecedent.

\paragraph{Syntactic features.}

The syntactic features are based on running an in-house state of the
art part of speech tagger and syntactic chunker on the data.  The
words include unigrams and bigrams of part of speech as well as
unigram chunk features.  We have not used any parsing for this task.

\paragraph{Pattern-based features.}

We have included a whole slew of features based on lexical and part of
speech patterns surrounding the current word.  These include: eight
hand-written patterns for identifying pleonastic ``it'' and ``that''
(as in ``It is raining'' or ``It seems to be the case that \dots'');
identification of pluralization features on the previous and next head
nouns (this is intended to help make decisions about entity types);
the previous and next content verb (also intended to help with entity
type identification); the possessor or possessee in the case of simple
possessive constructions (``The president 's speech'' would yield a
feature of ``president'' on the word ``speech'', and vice-versa; this
is indented to be a sort of weak sub-categorization principle); a
similar feature but applied to the previous and next content verbs
(again to provide a weak sort of sub-categorization); and, for
coreference, a list of part of speech and word sequence patterns that
match up to four words between nearby mentions that are either highly
indicative of coreference (e.g., ``of,'' ``said,'' ``am'' ``, a'') or
highly indicative of non-coreference (e.g., ``'s,'' ``and,'' ``in
the,'' ``and the'').  This last set was generated by looking at
intervening strings and finding the top twenty that had maximal mutual
information with with class (coreferent or not coreferent) across the
training data.

\paragraph{Count-based features.}

The count-based features apply only to the coreference task and
attempt to capture regularities in the size and distribution of
coreference chains.  These include: the total number of entities
detected thus far; the total number of mentions; the entity to mention
ratio; the entity to word ratio; the mention to word ratio; the size
of the hypothesized entity chain; the ratio of the number of mentions
in the current entity chain to the total number of mentions; the
number of intervening mentions between the current mention and the
last one in our chain; the number of intervening mentions of the same
type; the number of intervening sentence breaks; the Hobbs distance
computed over syntactic chunks; and the ``decayed density'' of the
hypothesized entity, which is computed as $\sum_{m=e} 0.5^{d(m)} /
\sum_{m} 0.5^{d(m)}$, where $m$ ranges over all previous mentions
(constrained in the numerator to be in the same coreference chain as
our mention) and $d(m)$ is the number of entities away this mention
is.  This feature is captures that some entities are referred to
consistently across a document, while others are mentioned only for
short segments, but it is relatively rare for an entity to be
mentioned once at the beginning and then ignored again until the end.

\paragraph{Semantic features.}

The semantic features used are drawn from WordNet \cite{fellbaum98}.
They include: the two most common synsets from WordNet for all the
words in a chunk; all hypernyms of those synsets; for coreference,
we also consider the distance in the WordNet graph between pairs of
head words (defined to be the final word in the mention name) and
whether one is a part of the other.  Finally, we include the synset
and hypernym information of the preceding and following verbs, again
to model a sort of sub-categorization principle.

\paragraph{Knowledge-based features.}

Based on the hypothesis that many name to nominal coreference chains
are best understood in terms of background knowledge (for instance,
that ``George W. Bush'' is the ``President''), we have attempted to
take advantage of recent techniques from large scale data mining to
extract lists of such pairs.  In particular, we use the name/instance
lists described by \cite{fleischman-answering03} and available on
Fleischman's web page to generate features between names and nominals
(this list contains $2m$ pairs mined from $15$GBs of news data).
Since this data set tends to focus mostly on person instances from
news, we have additionally used similar data mined from a $138$GB web
corpus, for which more general ``ISA'' relations were mined
\cite{ravichandran05randomized}.

\paragraph{Class-based features.}

The class-based features we employ are designed to get around the
sparsity of data problem while simultaneously providing new
information about word usage.  The first class-based feature we use is
based on word classes derived from the web corpus mentioned earlier
and computed as described by \cite{ravichandran05randomized}.  The
second attempts to instill knowledge of collocations in the data; we
use the technique described by \cite{dunning93coincidence} to compute
multi-word expressions and then mark words that are commonly used as
such with a feature that expresses this fact.

\paragraph{List-based features.}

We have gathered a collection of about 40 lists of common places,
organization, names, etc.  These include the standard lists of names
gathered from census data and baby name books, as well as standard
gazetteer information listing countries, cities, islands, ports,
provinces and states.  We supplement these standard lists with lists
of airport locations (gathered from the FAA) and company names (mined
from the NASDAQ and NYSE web pages).  We additionally include lists of
semantically plural but syntactically singular words (e.g., ``group'')
which were mined from a large corpus by looking for patterns such as
(``members of the \dots'').  Finally, we use a list of persons,
organizations and locations that were identified at least 100 times in
a large corpus by the BBN IdentiFinder named entity tagger
\cite{bikel99name}.

These lists are used in three ways.  First, we use simple list
membership as a feature to improve detection performance.  Second, for
coreference, we look for word pairs that appear on the same list but
are not identical (for instance, ``Russia'' and ``England'' appearing
on the ``country'' list but not being identical hints that they are
different entities).  Finally, we look for pairs where one element in
the pair is the head word from one mention and the other element in
the pair is a list.  This is intended to capture the notion that a
word that appears on out ``country list'' is often coreferent with the
word ``country.''

\paragraph{Inference-based features.}

The inference-based features are computed by attempting to infer an
underlying semantic property of a given mention.  In particular, we
attempt to identify gender and semantic number (e.g., ``group'' is
semantically plural although it is syntactically singular).  To do so,
we created a corpus of example mentions labels with number and
gender, respectively.  This data set was automatically extracted from
our EDT data set by looking for words that corefer with pronouns for
which we know the number or gender.  For instance, a mention that
corefers with ``she'' is known to be singular and female, while a
mention that corefers with ``they'' is known to be plural.  In about
5\% of the cases, this was ambiguous -- these cases were thrown out.
We then used essentially the same features as described above to build
a maximum entropy model for predicting number and gender.  The
predictions of this model are used both as features for detection as
well as coreference (in the latter case, we check for matches).
Additionally, we use several pre-existing classifiers as features.
This are simple maximum entropy Markov models trained off of the MUC6
data, the MUC7 data and our ACE data.

\paragraph{String match features.}

We use the standard string match features that are described in every
other coreference paper.  These are: string match; substring match;
string overlap; pronoun match; and normalized edit distance.  In
addition, we also use a string nationality match, which matches, for
instance ``Israel'' and ``Israeli,'' ``Russia'' and ``Russian,''
``England'' and ``English,'' but not ``Netherlands'' and ``Dutch.''
This is done by checking for common suffixes on nationalities and
matching the first half of the of the words based on exact match.  We
additionally use a linguistically-motivated string edit distance,
where the replacement costs are lower for vowels and other easily
confusable characters.  We also use the Jaro distance as an additional
string distance metric.  Finally, we attempt to match acronyms by
looking at initial letters from the words in long chunks.

\paragraph{History-based features.}

Finally, for the detection phase of the task, we include features
having to do with long-range dependencies between words.  For
instance, if at the beginning of the document we tagged the word
``Arafat'' as a person's name (perhaps because it followed ``Mr.'' or
``Palestinian leader''), and later in the document we again see the
word ``Arafat,'' we should be more likely to call this a person's
name, again.  Such features have previously been explored in the
context of information extraction from meeting announcements using
conditional random fields augmented with long-range links
\cite{sutton04distant}, but the LaSO framework makes no Markov
assumption, so there is no extra effort required to include such
features.

\vspace{-1.5mm}
\subsection{Decision Features}

Our decision features are divided into three classes: simple,
coreference and boundary features.

\paragraph{Simple.}

The simple decision features include: is this chunk tagged as an
entity; what is its entity type; what is its entity subtype; what is
its mention type; what is its entity type/mention type pair.

\paragraph{Coreference.}

The coreference decision features include: is this entity the start of
a chain or continuing an existing chain; what is the entity type of
this started (or continued) chain; what is the entity subtype of this
started (or continued) chain; what is the mention type of this started
chain; what is the mention type of this continued chain and the
mention type of the most recent antecedent.

\paragraph{Boundary.}

The boundary decision features include: the second and third order
Markov features over entity type, entity subtype and mention type;
features appearing at the previous (and next) words within a window of
three; the words that appear and the previous and next mention
boundaries, specified also by entity type, entity subtype and mention
type.

\vspace{-1.5mm}\section{Experimental Results} \label{sec:results}\vspace{-1.5mm}

\vspace{-1.5mm}
\subsection{Data}

We use the official 2004 ACE training and test set for evaluation
purposes; however, we exclude from the training set the Fisher
conversations data, since this is very different from the other data
sets and there is no Fisher data in the 2004 test set.  This amounts
to $392$ training documents, consisting of $8.1k$ sentences and $160k$
words.  There are a total of $24k$ mentions in the data corresponding
to $10k$ entities (note that the data is not annotated for
cross-document coreference, so instances of ``Bill Clinton'' appearing
in two different documents are counted as two different entities).
Roughly half of the entities are people, a fifth are organizations, a
fifth are GPEs and the remaining are mostly locations or facilities.
The test data is $192$ documents, $3.5k$ sentences and $64k$ words,
with $10k$ mentions to $4.5k$ entities.  In all cases, we use a beam
of 16 for training and test, and ignore features that occur fewer than
five times in the training data.

\vspace{-1.5mm}
\subsection{Evaluation Metrics}

There are many evaluation metrics possible for this data.  We will use
as our primary measure of quality the ACE metric.  This is computed,
roughly, by first matching system mentions with reference mentions,
then using those to match system entities with reference entities.
There are costs, once this matching is complete, for type errors,
false alarms and misses, which are combined together to give an ACE
score, ranging from $0$ to $100$, with $100$ being perfect (we use
v.10 of the ACE evaluation script).

\vspace{-1.5mm}
\subsection{Joint versus Pipelined}

We compare the performance of the joint system with the pipelined
system.  For the pipelined system, to build the mention detection
module, we use the same technique as for the full system, but simply
don't include in the hypotheses the coreference chain information
(essentially treating each mention as if it were in its own chain).
For the stand-alone coreference system, we assume that the correct
mentions and types are always given, and simply hypothesize the chain
(though still in a left-to-right manner).\footnote{One subtle
difficulty with the joint model has to do with the online nature of
the learning algorithm: at the beginning of training, the model is
guessing randomly at what words are entities and what words are not
entities.  Because of the large number of initial errors made in this
part of the task, the weights learned by the coreference model are
initially very noisy.  We experimented with two methods for
compensating for this effect.  The first was to give the mention
identification model as ``head start'': it was run for one full pass
through the training data, ignoring the coreference aspect and the
following iterations were then trained jointly.  The second method was
to only update the coreference weights when the mention was identified
correctly.  On development data, the second was more efficient and
outperformed the first by $0.6$ ACE score, so we use this for the
experiments reported in this section.}  Run as such, the joint model
achieves an ACE score of $79.4$ and the pipelined model achieves an
ACE score of $78.1$, a reasonably substantial improvement for
performing both task simultaneously.  We have also computed the
performance of these two systems, ignoring the coreference scores
(this is done by considering each mention to be its own entity and
recomputing the ACE score).  In this case, the joint model, ignoring
its coreference output, achieves an ACE score of $85.6$ and the
pipelined model achieves a score of $85.3$.  The joint model does
marginally better, but it is unlikely to be statistically significant.
In the 2004 ACE evaluation, the best three performing systems achieved
scores of $79.9$, $79.7$ and $78.2$; it is unlikely that our system is
significantly worse than these.

\vspace{-1.5mm}
\subsection{Feature Comparison for Coreference}

\begin{figure}[t]
\center
\psfig{figure=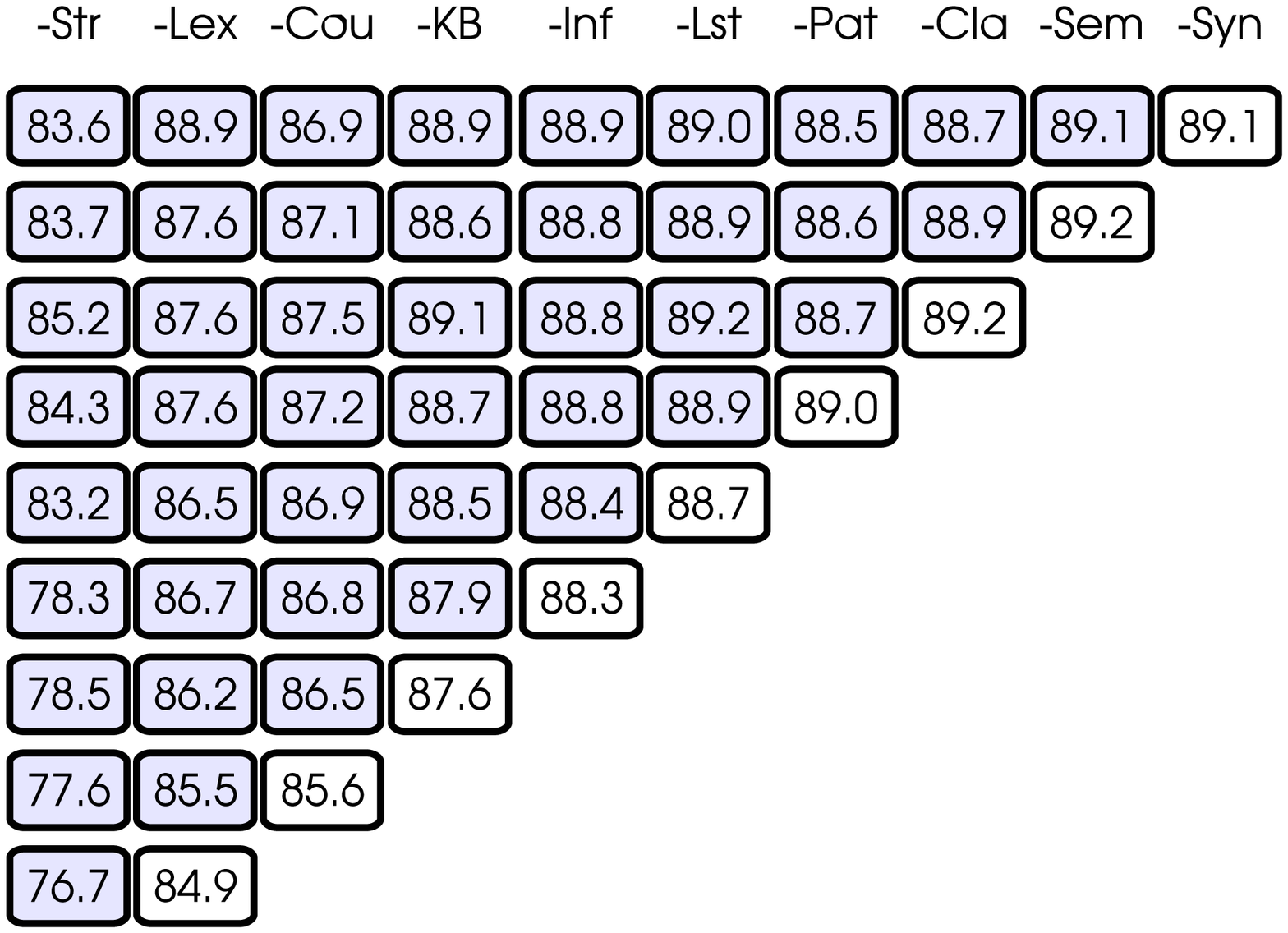,width=8cm}
\caption{Comparison of performance as different feature classes are
removed.}
\label{fig:feature-sel}
\end{figure}

In this section, we analyze the effects of the different base feature
types on coreference performance.  We use a model with perfect
mentions, entity types and mention types (with the exception of
pronouns: we do not assume we know pronoun types, since this gives
away too much information), and measure the performance of the
coreference system.  When run with the full feature set, the model
achieves an ACE score of $89.1$ and when run with no added features
beyond simple biases, it achieves $65.4$.  The best performing system
in the 2004 ACE competition achieved a score of $91.5$ on this task;
the next best system scored $88.2$, which puts us squarely in the
middle of these two (though, likely not statistically significantly
different).  Moreover, the best performing system took advantage of
additional data that they labeled in house.

To compute feature performance, we begin with all feature types and
iteratively remove them one-by-one so that we get the best performance
(we do not include the ``history'' features, since these are not
relevant to the coreference task).  The results are shown in
Figure~\ref{fig:feature-sel}.  Across the top line, we list the ten
feature classes.  The first row of results shows the performance of
the system after removing just one feature class.  In this case,
removing lexical features reduces performance to $88.9$, while
removing string-match features reduces performance to $83.6$.  The
non-shaded box (in this case, syntactic features) shows the feature
set that can be removed with the least penalty in performance.  The
second row repeats this, after removing syntactic features.

As we can see from this figure, we can freely remove syntax, semantics
and classes with little decrease in performance.  From that point,
patterns are dropped, followed by lists and inference, each with a
performance drop of about $0.4$ or $0.5$.  Removing the knowledge
based features results in a large drop from $87.6$ down to $85.6$ and
removing count-based features drops the performance another $0.7$
points.  Based on this, we can easily conclude that the most important
feature classes to the coreference problem are, in order, string
matching features, lexical features, count features and
knowledge-based features, the latter two of which are novel to this
work.

\vspace{-1.5mm}
\subsection{Linkage Types}

As stated in the previous section, the coreference-only task with
intelligent link achieves an ACE score of $89.1$.  The next best score
is with min link ($88.7$) followed by average link with a score of
$88.1$.  There is then a rather large drop with max link to $86.2$,
followed by another drop for last link to $83.5$ and first link
performs the poorest, scoring $81.5$.

\vspace{-1.5mm}\section{Discussion} \label{sec:discussion}\vspace{-1.5mm}

In this paper, we have applied the \emph{Learning as Search
Optimization (LaSO)} framework to the entity detection and tracking
task.  The framework is an excellent choice for this problem, due to
the fact that many relevant features for the coreference task (and
even for the mention detection task) are highly non-local.  This
non-locality makes models like Markov networks intractable, and LaSO
provides an excellent framework for tackling this problem.  We have
introduced a large set of new, useful features for this task, most
specifically the use of knowledge-based features for helping with the
name-to-nominal problem, which has led to a substantial improvement in
performance.  We have shown that performing joint learning for mention
detection and coreference results in a better performing model that
pipelined learning.  We have also provided a comparison of the
contributions of our various feature classes and compared different
linkage types for coreference chains.  In the process, we have
developed an efficient model that is competitive with the best ACE
systems.

Despite these successes, our model is not perfect: the largest source
of error is with pronouns.  This is masked by the fact that the ACE
metric weights pronouns low, but a solution to the EDT problem should
handle pronouns well.  We intend to explore more complex features for
resolving pronouns, and to incorporate these features into our current
model.  We also intend to explore more complex models for
automatically extracting knowledge from data that can help with this
task and applying this technique to a real application, such as
summarization.

\begin{small}
\paragraph{Acknowledgments:}
We thank three anonymous reviewers for helpful comments.  This work
was supported by DARPA-ITO grant NN66001-00-1-9814 and NSF grant
IIS-0326276.
\end{small}



\end{document}